\documentclass{svproc}

\usepackage{url}
\usepackage{hyperref}
\hypersetup{
    colorlinks=true,      
    linkcolor=blue,       
    citecolor=green,      
    filecolor=magenta,    
    urlcolor=red          
}

\usepackage{graphicx}
\usepackage{float}
\usepackage{amsmath}
\usepackage{tabularx}
\usepackage{amsmath,amssymb,amsfonts}
\usepackage{lipsum}  
\usepackage{cite}
\usepackage{algorithmic}
\usepackage{textcomp}
\usepackage{xcolor}
\usepackage{booktabs}

\begin{document}
\mainmatter    
\title{BLM-SGAN: Bidirectional Language Modeling for Semantic-Spatial Text-to-Image Generation}
\titlerunning{BLM-SGAN}  
%
\author{
    Ahmed Abdelmoneim Mazrou \and
    Haidy Maher El-Amir \and
    Ali Hamdi
}
\institute{
    Faculty of Computer Science, MSA University, Egypt \\
\texttt{{\{ahmed.abdelmoneim2, haidy.maher, ahamdi\}}@msa.edu.eg}
}

\maketitle      

\begin{abstract}
Despite the success of image generation from text descriptions, it still faces challenges that are difficult to overcome in domains such as natural language processing (NLP) and computer vision (CV). Recent advancements in text-to-image (T2I) models, particularly those utilizing generative adversarial networks (GANs), have significantly improved the synthesis of realistic images across various domains. However, existing GAN-based T2I models still encounter key challenges, such as difficulty in capturing long-range dependencies, vanishing gradients, and the limitations of sequential processing. To address these issues, we introduce BLM-SGAN—a novel model that incorporates Bidirectional Language Modeling for Semantic-Spatial Text-to-Image Generation. BLM-SGAN leverages BERT's attention mechanisms to capture rich contextual information and efficiently manage extended sequences. Our model demonstrates state-of-the-art performance, with an Inception Score (IS) of 5.45 ± 0.08, surpassing several competitive models such as SSA-GAN, DF-GAN, SD-GAN, and AttnGAN. BLM-SGAN effectively generates highly realistic images of birds from detailed text descriptions. The implementation code is available at: \href{https://github.com/haidy-maher/BLM-SGAN-Text-to-Image-Generation}{https://github.com/haidy-maher/BLM-SGAN-Text-to-Image-Generation}.

\end{abstract}

\section{Introduction}

In recent times, image synthesis has evolved into a widely popular topic due to its remarkable capabilities. One of the most prominent and rapidly advancing approaches in this field is Generative Adversarial Networks (GANs) \cite{goodfellow2014generative,kassab2024mmis}, which have risen as a significant research topic due to their wide range of applications, including photo editing and computer-aided design. One particularly important application of GANs is text-to-image (T2I) generation. T2I bridges the gap between two distinct domains—natural language processing and computer vision—allowing linguistic descriptions to guide image generation. This approach resonates with how people naturally communicate and describe visual scenes, making it highly intuitive. However, T2I generation is challenging due to the cross-modal nature of the task, which requires generating images that are semantically consistent with the provided text descriptions \cite{liao2022text}.

Prior methodologies for T2I production have often utilized stacking multiple generator-discriminator pairs to enhance image quality \cite{zhang2021dtgan,zhang2018photographic,hong2018inferring,xu2018attngan,li2019controllable, yin2019semantics}. While this technique has proven effective, it demands significant computational resources and results in an unstable training process. Furthermore, the quality of images generated by early stages influences later stages; poor-quality images produced by the initial generators limit the performance of subsequent ones. To overcome these challenges, we adopt a one-stage architecture \cite{liao2022text, tao2022df}, which simplifies the process while maintaining high image fidelity.

\begin{figure}[H]
    \centering
    {\small \textbf{Input: A bright \{color\}{} bird with dark lines, wings, and tiny eyes.}} 
    \vspace{0.5em}  
    \begin{minipage}[b]{0.25\linewidth}
        \centering
        {\small \textbf{yellow}}  
        \includegraphics[width=\linewidth]{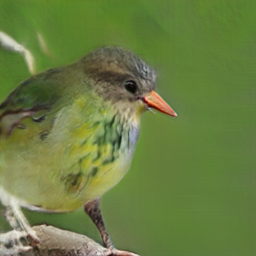}
    \end{minipage}%
    \begin{minipage}[b]{0.25\linewidth}
        \centering
        {\small \textbf{white}}  
        \includegraphics[width=\linewidth]{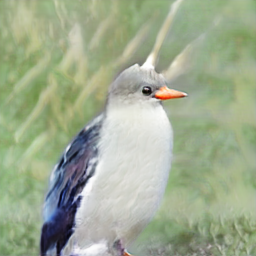}
    \end{minipage}%
    \begin{minipage}[b]{0.25\linewidth}
        \centering
        {\small \textbf{red}}  
        \includegraphics[width=\linewidth]{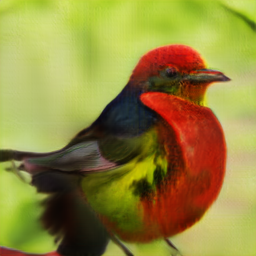}
    \end{minipage}%
    \begin{minipage}[b]{0.25\linewidth}
        \centering
        {\small \textbf{blue}}  
        \includegraphics[width=\linewidth]{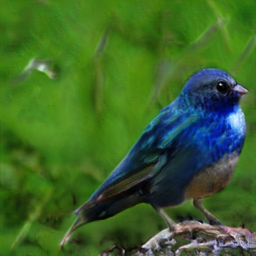}
    \end{minipage}
    
    \caption{Bird images generated from different color text inputs.}
    \label{samples}
\end{figure}

Moreover, current T2I models struggle to fuse textual and visual information effectively. The main methods used for text-image information fusion include feature concatenation, cross-modal attention, and Conditional Batch Normalization (CBN). Previous studies \cite{reed2016generative, zhang2017stackgan, zhang2018stackgan++} employed basic concatenation techniques, which failed to fully leverage the text data or achieve deep fusion between text and image features. Cross-modal attention is used by recent models like AttnGAN \cite{xu2018attngan} to determine word-context vectors for each sub-region in the image, yet it results in an increase in computational cost as the image size increases. The abstract, high-level semantics conveyed in natural language often mismatch the low-level details present in images \cite{chen2017sca, yu2017multi}, making it difficult to precisely control image generation, Particularly for intricate images containing multiple objects.

In addition, models such as SD-GAN \cite{yin2019semantics} have implemented word-level and sentence-level Conditional Batch Normalizations to integrate textual information into image feature maps. Still, this approach applies the text features only a few times during generation, limiting the depth of text-image fusion. DF-GAN \cite{tao2022df} addresses this by applying stacked affine transformations, where parameters derived from the text vector scale and shift image features channel-wise. However, this technique does not account for the spatial relevance of the text, applying transformations uniformly across the image. SSA-GAN \cite{liao2022text} further improves text-image fusion by introducing Semantic-Spatial Attention Convolutional Network (SSACN) blocks that ensure text features are selectively injected into the most relevant sub-regions of the image. This guarantees a more precise and context-aware fusion of text and image features throughout the generation process, addressing the limitations of uniform transformations in earlier models. However, despite these advancements, the challenge of fully integrating deep contextual information from the text, while maintaining a balance between semantic relevance and spatial precision, remains. Addressing this challenge, we introduce BLM-SGAN, a novel model that integrates Bidirectional Language Modeling to enhance text-image fusion and improve the quality of generated images.

\begin{figure*}[t]
    \centering
    \makebox[\textwidth][c]{%
        \includegraphics[width=1.1\textwidth]{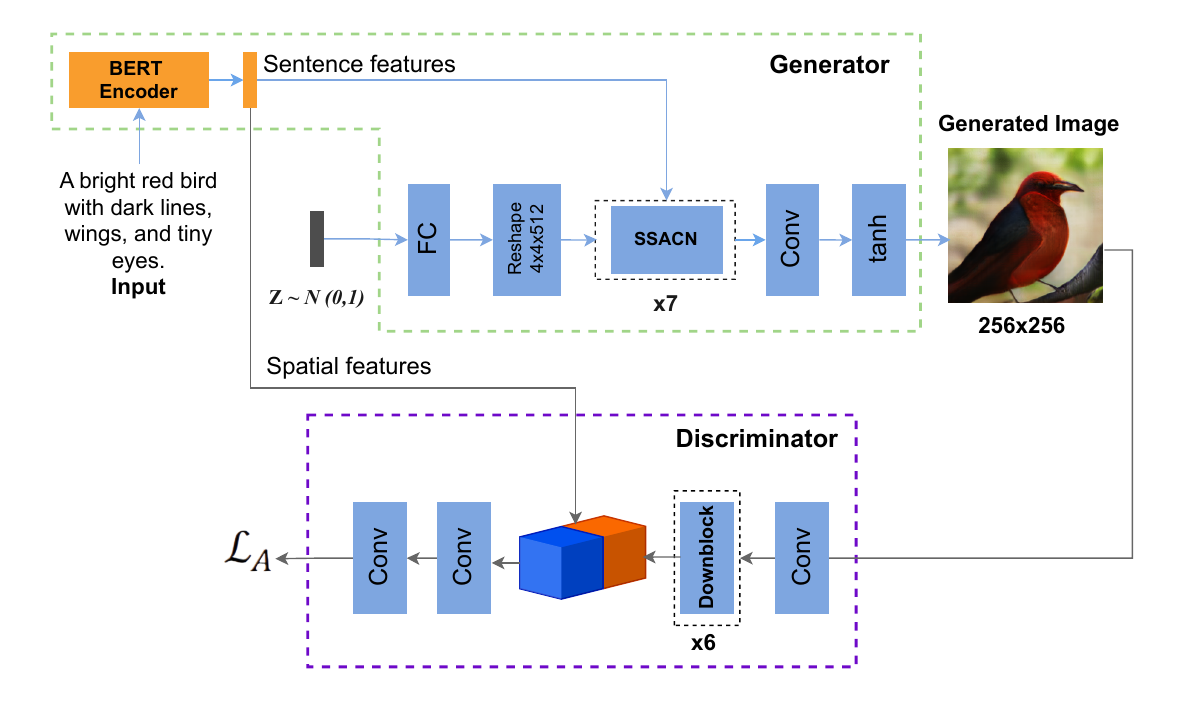}
    }
    \caption{
    A schematic of our BLM-SGAN framework, featuring a generator-discriminator pair. The generator includes 7 SSACN blocks that integrate text with image features to ensure semantic consistency. BERT is used as the text encoder, enhancing contextual understanding. Gray lines indicate training-only data flows.}
    \label{figure:ex}
\end{figure*}

Our proposed model, \textbf{BLM-SGAN}, overcomes these limitations by integrating Bidirectional Language Modeling (BLM) into the Semantic-Spatial Text-to-Image generation process. As illustrated in Figure \ref{figure:ex}, BLM-SGAN leverages BERT's powerful attention mechanisms to capture rich contextual information, ensuring that text features are effectively mapped to the most relevant sub-regions of the image throughout the generation process.

This paper presents BLM-SGAN as a novel solution to the challenges faced by current T2I models. Section 2 covers related work, detailing the strengths and limitations of existing approaches. Section 3 introduces our research methodology, focusing on how we integrate bidirectional language modeling into the image generation process. Section 4 presents our experimental results and discusses the effectiveness of our approach. Finally, Section 5 concludes with a summary of our contributions and directions for future research. 

\section{Related Work}

Text-to-image (T2I) generation, a critical task at the intersection of computer vision (CV) and natural language processing (NLP), has garnered significant attention in recent years. Generative Adversarial Networks (GANs) \cite{goodfellow2014generative} have become dominant for T2I generation, with conditional GANs (cGANs) \cite{yu2017unsupervised} first introduced by Reed et al. \cite{reed2016generative} as a way to generate images based on text descriptions. This approach bridges the gap between linguistic descriptions and visual representations, though it faces challenges in translating high-level semantic information into accurate, coherent images. 

Early T2I models like StackGAN \cite{zhang2017stackgan,zhang2018stackgan++} adopted a multi-stage architecture to progressively refine images. Each generator in the stack was paired with a discriminator to enhance image quality through adversarial learning. However, this multi-stage approach is computationally expensive and prone to training instability, particularly when initial generators produce low-quality outputs that hinder the refinement process in later stages. Zhu et al. \cite{zhu2019dm} addressed these issues with a dynamic memory module that iteratively enhances the initial images, improving the robustness of the stacked architecture. In contrast, more recent models have shifted toward one-stage designs to simplify the pipeline and reduce computational costs. For instance, Ming et al. \cite{tao2022df} proposed DF-GAN, which employs a single generator-discriminator pair and incorporates UPBlocks for upsampling to produce high-resolution images more efficiently. These one-stage models mitigate the cumulative error problem of multi-stage architectures while maintaining high-quality results. Our work follows this trend by adopting a one-stage approach to streamline the generation mechanism.

A major challenge in T2I generation is the effective fusion of text and image information. Early models \cite{reed2016generative,zhang2017stackgan,zhang2018stackgan++} typically concatenated text embeddings with visual feature maps or noise vectors, leading to limited fusion between modalities. AttnGAN \cite{xu2018attngan} generates word-context vectors for sub-regions of an image using cross-modal attention mechanisms, thereby improving the fusion of text and images. Additionally, the Deep Attentional Multimodal Similarity Model (DAMSM) is used to assess the similarity between text and images, ensuring that the generated images are more semantically consistent. Building on this, ControlGAN \cite{li2019controllable} applied spatial and channel-wise attention blocks to improve the granularity of text-image fusion. 

Other models, such as DM-GAN \cite{zhu2019dm}, utilized memory networks to dynamically refine image features based on important textual information, while SD-GAN \cite{Yin_2019_CVPR} introduced semantic-conditioned batch normalization (BN) to integrate global sentence and word vectors into the image generation process. DF-GAN \cite{tao2022df} further refined this idea by applying stacked affine adjustments guided by the text embeddings to modify image features throughout the generation process. The Transformer model introduced by Vaswani et al. \cite{vaswani2017attention} in "Attention is All You Need" utilizes self-attention mechanisms to capture long-range dependencies in text, enhancing the semantic understanding necessary for accurate image synthesis. Building on this, BERT (Bidirectional Encoder Representations from Transformers) \cite{devlin2018bert} employs a bidirectional approach to generate rich, context-aware embeddings.

Our proposed model, \textbf{BLM-SGAN}, builds on these advancements by introducing Bidirectional Language Modeling (BLM) for Semantic-Spatial Text-to-Image Generation. Leveraging BERT's powerful attention mechanisms, BLM-SGAN enhances text representation and enables more effective fusion of text and image features. These embeddings of BERT improve the alignment between textual descriptions and visual outputs. Additionally, BERT’s bidirectional language modeling capabilities allow BLM-SGAN to capture richer contextual information compared to traditional LSTM or GRU-based models, improving the overall accuracy and consistency of the generated images. By integrating BERT into the T2I generation process, our model better bridges the gap between textual descriptions and visual outputs, enabling more accurate and context-aware image generation.

\section{Methodology}
The architecture of our proposed model, BLM-SGAN (Bidirectional Language Modeling for Semantic-Spatial Text-to-Image Generation), is depicted in Fig.\ref{figure:ex}. Following the one-stage structure used in \cite{tao2022df, liao2022text}, we enhance the SSA-GAN architecture \cite{liao2022text} by replacing the LSTM-based text encoder with a BERT transformer. Our model comprises three main components: (1) a BERT-based text encoder for richer text representation, (2) a generator with 7 SSACN blocks that improve text-image fusion and resolution, and (3) a discriminator 
that ensures semantic consistency between the generated images and the input text.

The model takes as input a text description and a 100-dimensional noise vector sampled from a Gaussian distribution, and outputs a 256 by 256 pixel RGB image. We elaborate on each component below.

\subsection{Text Encoder}
Our model integrates BERT (Bidirectional Encoder Representations from Transformers) as the primary text encoder, enhancing performance over traditional bidirectional LSTM networks \cite{schuster1997bidirectional,xu2018attngan,zhu2019dm}. This approach leverages the advancements in transformer-based architectures, especially those introduced in "Attention is All You Need" \cite{vaswani2017attention}, which laid the foundation for BERT \cite{devlin2018bert}.

BERT's bidirectional context processing allows it to interpret each word based on its full context—both preceding and following words—unlike the sequential approach in LSTMs. This bidirectionality enables BERT to generate rich contextual embeddings, essential for converting detailed textual descriptions into semantically aligned visual elements within generated images. As shown in Fig.~\ref{figure:ex}, BERT generates sentence features that are passed into two main parts: the SSACN blocks within the generator and the discriminator via spatial replication.

To define the encoding mathematically, BERT processes each input as a token sequence \( \mathbf{X} = \{x_1, x_2, \dots, x_n\} \), where each token \( x_i \) is transformed into an embedding. The model’s stacked transformer layers, each containing self-attention and feed-forward neural components, enable tokens to assess the importance of all others in the sequence, capturing long-range dependencies essential for a deep contextual understanding.

The attention mechanism, calculated as:

\[
\text{Attention}(Q, K, V) = \text{softmax}\left(\frac{QK^T}{\sqrt{d_k}}\right)V
\]

includes query, key, and value matrices extracted from token embeddings, with the key dimensionality being a critical factor. This enables BERT to weigh relationships within the sequence, creating contextual representations that enrich semantic alignment.

Each transformer layer outputs a contextualized embedding for each token, represented as \( \mathbf{H}_i = \text{TransformerLayer}(\mathbf{H}_{i-1}) \) for \( i = 1, 2, \dots, L \), where \( L \) is the total number of layers. The final output \( \mathbf{H}_L \) provides bidirectional representations for all tokens, setting the stage for robust text-image fusion.

BERT’s effectiveness stems from its pre-training tasks: Masked Language Modeling (MLM) and Next Sentence Prediction (NSP). MLM enables BERT to understand context by predicting masked words, while NSP helps capture sentence relationships. The encoded output \( \mathbf{H} = \{h_1, h_2, \dots, h_n\} \), with each \( h_i \) representing a token embedding, is split into two paths: one feeding into the SSACN blocks within the generator for enhancing text-image fusion, and the other spatially replicated for use in the discriminator to ensure semantic consistency (see Fig.~\ref{figure:ex}). Spatial replication allows the sentence features to be aligned with each spatial location in the generated image, providing the discriminator with localized contextual information for verifying semantic alignment across the entire image. By embedding BERT’s attention mechanisms, we provide text features that guide the generation process, enabling realistic, contextually precise images.

\subsection{Semantic-Spatial Aware Convolutional Network (SSACN)}
The generator in BLM-SGAN leverages the Semantic-Spatial Aware Convolutional Network (SSACN) introduced by \cite{liao2022text}, which fuses the encoded text features with image feature maps at each generation stage. The SSACN block receives the BERT encoded text vector $\bar{e}$ and image feature maps from the previous block, producing refined image feature maps from the previous block, producing refined image feature maps represented as a three-dimensional tensor with dimensions corresponding to channel count, height, and width. The first SSACN block projects the noise vector \( z \) into the visual domain, resulting in a $4 \times 4 \times 512$ feature map. Subsequent blocks progressively upsample the image feature maps, reaching the final resolution of $256 \times 256$. The main components of the SSACN block are the mask predictor, upsampling module, Semantic-Spatial Conditional Batch Normalization (SSCBN) layer, and residual block. The mask predictor selectively applies text information to relevant regions of the image feature maps, improving semantic consistency without overwriting unrelated details. This helps preserve image quality while integrating text-relevant features effectively.

\subsubsection{Mask Generator Under Weak Guided Supervision}

 shown within the black dashed box in Fig.\ref{figure:ex}, is a core component of the SSACN block. It consists of a series of convolutional layers followed by ReLU activation and batch normalization, ending with a sigmoid layer to produce mask maps \( m_i \in \mathbb{R}^{h_i \times w_i} \) in the range [0, 1]. These masks help modulate the influence of textual information on different image regions, emphasizing areas that need alignment with the text. The mask \( m_i(h, w) \) from the predictor is applied to control the spatially adaptive affine transformation, adjusting the scale and shift of the feature maps based on both spatial locations and the enhanced text features from BERT. Trained jointly with the network and guided by the discriminator’s adversarial loss.

\subsection{Discriminator}
The discriminator in BLM-SGAN distinguishes between real and generated images, guiding the generator to produce more realistic and semantically accurate outputs. As shown in Fig. \ref{figure:ex} (violet dashed box), it follows the one-way design from \cite{tao2022df, liao2022text}, ensuring simplicity and effectiveness. The discriminator concatenates the generated image features with the BERT-encoded text vector to compute the adversarial loss via convolutional layers. To ensure semantically consistent images, we apply a regularization technique that penalizes the discriminator’s gradients when they deviate from zero, maintaining smooth decision boundaries. The matching-aware component of this penalty ensures better alignment when generated image features closely match the text, improving consistency between the modalities. This drives the generator towards more accurate outputs \cite{tao2022df}. Additionally, a word-level fine-grained matching loss is applied using the Deep Attentional Multimodal Similarity Model (DAMSM) \cite{xu2018attngan}, preserving fine-grained textual details to enhance image quality.

\subsection{Loss Functions}
\textbf{Discriminator Loss Function:} The adversarial loss and MA-GP loss are combined for the training of the discriminator \cite{liao2022text}:
\begin{align*}
\mathcal{L}_{adv}^{D} = & \, \mathbb{E}_{x \sim p_{\text{data}}} \big[ \max(0, 1 - D(x, t)) \big] \\
& + 0.5 \, \mathbb{E}_{x \sim p_G} \big[ \max(0, 1 + D(\hat{x}, t)) \big] \\
& + 0.5 \, \mathbb{E}_{x \sim p_{\text{data}}} \big[ \max(0, 1 + D(x, \hat{t})) \big] \\
& + \lambda_M \mathbb{E}_{x \sim p_{\text{data}}} \Big[ \big( \| \nabla_x D(x, t) \|_2 + \| \nabla_t D(x, t) \|_2 \big)^p \Big].
\end{align*}

where \( t \) is the correct text description, \( \hat{t} \) is a mismatched text, \( x \) is the real image, and \( \hat{x} \) is the generated image. \( D(\cdot) \) is the discriminator output, which indicates whether the image-text pair is a match. The variables \( p \) and \( \lambda_M \) are hyperparameters for the MA-GP loss.

\textbf{Generator Loss Function:} In \cite{liao2022text}, a similar approach is employed where the generator's loss function combines both the adversarial loss and the DAMSM loss.
\[
\mathcal{L}_G = \mathcal{L}_{adv}^{G} + \lambda_{DA} \mathcal{L}_{DAMSM},
\]
\[
\mathcal{L}_{adv}^{G} = - \mathbb{E}_{x \sim p_G} \left[ D(\hat{x}, t) \right],
\]
\( \mathcal{L}_{DAMSM} \) denotes a word-level, fine-grained image-text matching loss, while \( \lambda_{DA} \) represents a weight for the DAMSM loss.

\section{Experimental Design}
This section provides an overview of the datasets, evaluation metrics, and implementation details applied in our experiments.

\subsection{Datasets}
The CUB (Caltech-UCSD Birds-200) dataset \cite{wah2011caltech} is used to evaluate our proposed BLM-SGAN model. The dataset includes a total of 11,788 images across 200 bird species, divided into 5,994 images for training and 5,794 images for testing. Each image is annotated with 10 textual descriptions, providing a detailed and varied set of input descriptions for the model. The diversity of bird species and the complexity of the descriptions make this dataset a rigorous tested for evaluating text-to-image synthesis models.

\subsection{Evaluation Metrics}

In this study, we adopt the Inception Score (IS) for evaluating the quality of generated images, a common metric for GANs \cite{salimans2016improved}. Higher IS values indicate that the images not only exhibit higher quality but also demonstrate more diversity and are more accurately aligned with their respective class labels. Kullback-Leibler (KL) divergence is computed using the Inception v3 network \cite{szegedy2016rethinking} to calculate the Inception Score (IS). This KL-divergence is measured between the distributions of generated images and real images.

\subsection{Implementation Details}
The following table summarizes the key implementation details of the model:

\begin{table}[H]
\centering
\caption{Implementation details of the model.}
\label{tab:implementation_details}
\begin{tabular}{ll}
\toprule
\textbf{Implementation Details}      & \textbf{Values} \\ 
\midrule
\textbf{Framework}                   & PyTorch \\ 
\textbf{Platform}                    & Google Colab Pro (Tesla T4 GPUs) \\ 
\textbf{Batch Size}                  & 8 \\ 
\textbf{Optimizer}                   & Adam \cite{kingma2014adam} \\ 
\textbf{Adam Parameters}             & $\beta_1 = 0.0$, $\beta_2 = 0.9$ \\ 
\textbf{Learning Rates (Generator)}  & 0.0001 \\ 
\textbf{Learning Rates (Discriminator)} & 0.0004 \\ 
\textbf{Hyperparameters}             & \( p = 6, \lambda_{MA} = 2, \lambda_{DA} = 0.1 \) \\ 
\textbf{Training Epochs}             & 156 \\ 
\textbf{Dataset}                     & CUB \\ 
\bottomrule
\end{tabular}
\end{table}

\section{Results and Discussion}

We evaluate our proposed BLM-SGAN model on the CUB (Caltech-UCSD Birds-200) dataset \cite{wah2011caltech}, comparing it to state-of-the-art Text-to-Image (T2I) GANs as presented in \cite{liao2022text}. The Inception Score (IS) \cite{salimans2016improved} was used for evaluation.

We present quantitative results demonstrating that our model achieves superior IS scores compared to existing methods, even with significantly fewer training epochs. Qualitative results are also provided to showcase the high quality and semantic consistency of images generated by our BLM-SGAN model.

\subsection{Quantitative Results}
Table \ref{tab:my_label} presents the performance of different models in terms of the Inception Score (IS). Our proposed BLM-SGAN model outperforms all baseline methods, achieving an IS score of \textbf{5.45 ± 0.08}. This result highlights the effectiveness
 of incorporating BERT-based bidirectional language modeling for generating
 higher-quality and semantically consistent images.

SSA-GAN \cite{liao2022text}, the closest competitor, achieves an IS score of 5.17 ± 0.08 after training for 600 epochs. Remarkably, our model surpasses SSA-GAN by achieving a higher IS score with only 156 epochs of training---a reduction of nearly 75\% in training time. This significant improvement demonstrates the efficiency of our approach. The ability to attain superior results with substantially fewer epochs not only indicates faster convergence but also reduces computational resources and training time. By effectively enhancing the fusion of text and image features through the Semantic-Spatial Aware Convolutional Network (SSACN) blocks and leveraging the advanced contextual understanding provided by BERT, our model delivers superior performance in a more efficient manner.

\begin{table}[H]
    \centering
    \caption{IS performance for various T2I models using the CUB dataset, with the highest results shown in bold.}
    \label{tab:my_label}
    \normalsize 
    \begin{tabularx}{0.6\textwidth}{X c c} \hline
         \textbf{Methods} & \textbf{IS $\uparrow$} \\ \hline 
         StackGAN++ \cite{zhang2018stackgan++} & 4.04 ± 0.06\\ 
         AttnGAN \cite{xu2018attngan} & 4.36 ± 0.03 \\ 
         ControlGAN \cite{li2019controllable} & 4.58 ± 0.09\\ 
         SD-GAN \cite{Yin_2019_CVPR} & 4.67 ± 0.09\\ 
         DM-GAN \cite{zhu2019dm} & 4.75 ± 0.07\\ 
         DF-GAN \cite{tao2022df} & 4.86 ± 0.04 \\ 
         SSA-GAN \cite{liao2022text} & 5.17 ± 0.08\\ \hline
         \textbf{Ours (BLM-SGAN)} & \textbf{5.45 ± 0.08} \\ \hline
    \end{tabularx}
\end{table}

\subsection{Qualitative Results}
Figure \ref{samples}
displays samples of generated images from our BLM-SGAN model, demonstrating the generation of high-resolution, semantically consistent images that accurately reflect the input textual descriptions. Compared to prior models, our approach produces finer details in bird features, such as beaks, feathers, and color patterns. Additionally, the images exhibit improved clarity and less blurring, which is particularly evident in the backgrounds.

\begin{figure}[H]
    \centering
    \begin{minipage}[b]{0.25\linewidth}
        \centering
        \includegraphics[width=\linewidth]{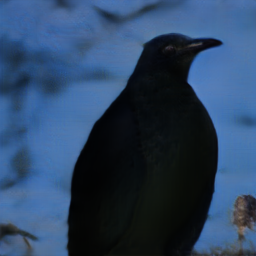}
    \end{minipage}%
    \begin{minipage}[b]{0.25\linewidth}
        \centering
        \includegraphics[width=\linewidth]{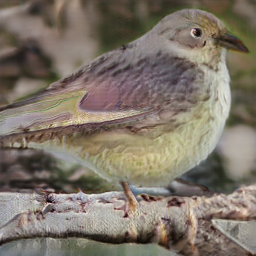}
    \end{minipage}%
    \begin{minipage}[b]{0.25\linewidth}
        \centering
        \includegraphics[width=\linewidth]{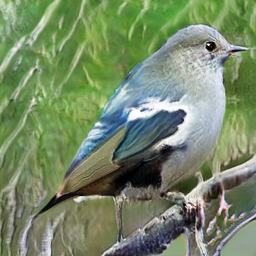}
    \end{minipage}%
    \begin{minipage}[b]{0.25\linewidth}
        \centering
        \includegraphics[width=\linewidth]{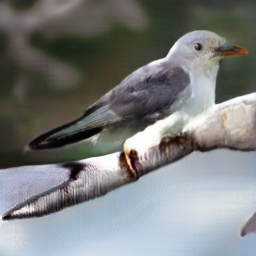}
    \end{minipage}
    
    \begin{minipage}[b]{0.25\linewidth}
        \centering
        \includegraphics[width=\linewidth]{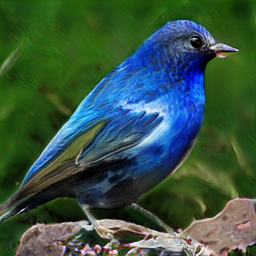}
    \end{minipage}%
    \begin{minipage}[b]{0.25\linewidth}
        \centering
        \includegraphics[width=\linewidth]{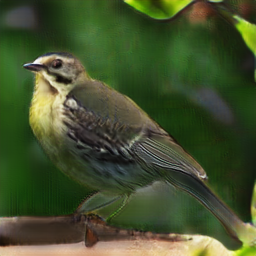}
    \end{minipage}%
    \begin{minipage}[b]{0.25\linewidth}
        \centering
        \includegraphics[width=\linewidth]{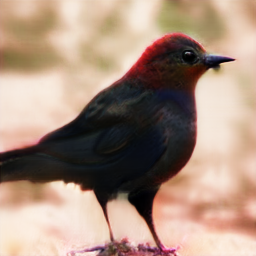}
    \end{minipage}%
    \begin{minipage}[b]{0.25\linewidth}
        \centering
        \includegraphics[width=\linewidth]{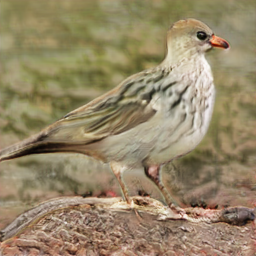}
    \end{minipage}
    \caption{Samples of generated bird images using BLM-SGAN.}
    \label{samples2}
\end{figure}

\section{Conclusion}
Within this study, we presented BLM-SGAN, an improved model for text-to-image (T2I) generation that builds upon and extends the SSA-GAN architecture. By integrating BERT with its bidirectional attention mechanism, our model captures richer contextual information from text descriptions and enhances the semantic alignment between text and image features. The introduction of Semantic-Spatial Aware Convolutional Networks (SSACN) further strengthens the fusion process, enabling precise application of text features to relevant image regions. Through Comprehensive experiments on the CUB dataset, we demonstrated that BLM-SGAN achieves better results than several state-of-the-art models, achieving superior quantitative results, as evidenced by higher Inception Scores, and generating more realistic and detailed images. These advancements contribute to bridging the gap between textual understanding and visual generation, pushing the boundaries of T2I generation. For future work, we will focus on testing BLM-SGAN with different datasets and explore how it performs with long, complex captions to further improve the model's ability to capture detailed semantic relationships.

\section{Acknowledgment}
We extend our heartfelt gratitude to AiTech AU, \textit{AiTech for Artificial Intelligence and Software Development} (\url{https://aitech.net.au}), for funding this research, providing technical support, and enabling its successful completion.

\bibliographystyle{bibtex/spmpsci}
\bibliography{ref}

@inproceedings{liao2022text,
  title={Text to image generation with semantic-spatial aware GAN},
  author={Liao, Wentong and Hu, Kai and Yang, Michael Ying and Rosenhahn, Bodo},
  booktitle={Proceedings of the IEEE/CVF Conference on Computer Vision and Pattern Recognition},
  pages={18187--18196},
  year={2022}
}

@inproceedings{goodfellow2014generative,
  title={Generative adversarial nets},
  author={Goodfellow, Ian and Pouget-Abadie, Jean and Mirza, Mehdi and others},
  booktitle={Advances in Neural Information Processing Systems (NeurIPS)},
  pages={2672--2680},
  year={2014}
}

@inproceedings{kassab2024mmis,
  title={MMIS: Multimodal Dataset for Interior Scene Visual Generation and Recognition},
  author={Kassab, Hozaifa and Mahmoud, Ahmed and Bahaa, Mohamed and Mohamed, Ammar and Hamdi, Ali},
  booktitle={2024 Intelligent Methods, Systems, and Applications (IMSA)},
  pages={172--177},
  year={2024},
  organization={IEEE}
}

@inproceedings{reed2016generative,
  title={Generative adversarial text to image synthesis},
  author={Reed, Scott and Akata, Zeynep and others},
  booktitle={International Conference on Machine Learning (ICML)},
  pages={1060--1069},
  year={2016}
}

@article{devlin2018bert,
  title={Bert: Pre-training of deep bidirectional transformers for language understanding},
  author={Devlin, Jacob},
  journal={arXiv preprint arXiv:1810.04805},
  year={2018}
}

@article{vaswani2017attention,
  title={Attention is all you need},
  author={Vaswani, A},
  journal={Advances in Neural Information Processing Systems},
  year={2017}
}

@inproceedings{yu2017multi,
  title={Multi-level attention networks for visual question answering},
  author={Yu, Dongfei and Fu, Jianlong and Mei, Tao and Rui, Yong},
  booktitle={Proceedings of the IEEE Conference on Computer Vision and Pattern Recognition},
  pages={4709--4717},
  year={2017}
}

@inproceedings{chen2017sca,
  title={Sca-cnn: Spatial and channel-wise attention in convolutional networks for image captioning},
  author={Chen, Long and Zhang, Hanwang and Xiao, Jun and others},
  booktitle={Proceedings of the IEEE Conference on Computer Vision and Pattern Recognition (CVPR)},
  pages={5659--5667},
  year={2017}
}

@inproceedings{hong2018inferring,
  title={Inferring semantic layout for hierarchical text-to-image synthesis},
  author={Hong, Seunghoon and Yang, Dingdong and Choi, Jongwook and Lee, Honglak},
  booktitle={Proceedings of the IEEE Conference on Computer Vision and Pattern Recognition (CVPR)},
  pages={7986--7994},
  year={2018}
}

@techreport{wah2011caltech,
  title={The Caltech-UCSD Birds-200-2011 Dataset},
  author={Wah, Catherine and Branson, Steve and Welinder, Peter and Perona, Pietro and Belongie, Serge},
  institution={California Institute of Technology},
  year={2011}
}

@inproceedings{yu2017unsupervised,
  title={Unsupervised representation learning with deep convolutional neural network for remote sensing images},
  author={Yu, Yang and Gong, Zhiqiang and Zhong, Ping and Shan, Jiaxin},
  booktitle={Image and Graphics: 9th International Conference, ICIG 2017, Shanghai, China, September 13-15, 2017, Revised Selected Papers, Part II 9},
  pages={97--108},
  year={2017},
  publisher={Springer}
}

@inproceedings{yin2019semantics,
  title={Semantics disentangling for text-to-image generation},
  author={Yin, Guojun and Liu, Bin and Sheng, Lu and others},
  booktitle={Proceedings of the IEEE/CVF Conference on Computer Vision and Pattern Recognition},
  pages={2327--2336},
  year={2019}
}

@inproceedings{tao2022df,
  title={DF-GAN: A simple and effective baseline for text-to-image synthesis},
  author={Tao, Ming and Tang, Hao and Wu, Fei and others},
  booktitle={Proceedings of the IEEE/CVF Conference on Computer Vision and Pattern Recognition},
  pages={16515--16525},
  year={2022}
}

@article{kingma2014adam,
  title={Adam: A method for stochastic optimization},
  author={Kingma, Diederik P and Ba, Jimmy},
  journal={arXiv preprint arXiv:1412.6980},
  year={2014}
}

@inproceedings{li2019controllable,
  title={Controllable text-to-image generation},
  author={Li, Bowen and Qi, Xiaojuan and Lukasiewicz, Thomas and Torr, Philip HS},
  booktitle={Advances in Neural Information Processing Systems (NeurIPS)},
  pages={2065--2075},
  year={2019}
}

@inproceedings{xu2018attngan,
  title={AttnGAN: Fine-grained text to image generation with attentional generative adversarial networks},
  author={Xu, Tao and Zhang, Pengchuan and Huang, Qiuyuan and others},
  booktitle={Proceedings of the IEEE Conference on Computer Vision and Pattern Recognition},
  pages={1316--1324},
  year={2018}
}

@inproceedings{zhang2018photographic,
  title={Photographic text-to-image synthesis with a hierarchically-nested adversarial network},
  author={Zhang, Zizhao and Xie, Yuanpu and Yang, Lin},
  booktitle={Proceedings of the IEEE Conference on Computer Vision and Pattern Recognition},
  pages={6199--6208},
  year={2018}
}

@inproceedings{zhang2021dtgan,
  title={DTGAN: Dual attention generative adversarial networks for text-to-image generation},
  author={Zhang, Zhenxing and Schomaker, Lambert},
  booktitle={2021 International Joint Conference on Neural Networks (IJCNN)},
  pages={1--8},
  year={2021}
}

@article{zhang2018stackgan++,
  title={StackGAN++: Realistic image synthesis with stacked generative adversarial networks},
  author={Zhang, Han and Xu, Tao and others},
  journal={IEEE Transactions on Pattern Analysis and Machine Intelligence},
  volume={41},
  number={8},
  pages={1947--1962},
  year={2018},
  publisher={IEEE}
}

@inproceedings{zhang2017stackgan,
  title={StackGAN: Text to photo-realistic image synthesis with stacked generative adversarial networks},
  author={Zhang, Han and Xu, Tao and others},
  booktitle={Proceedings of the IEEE/CVF International Conference on Computer Vision (ICCV)},
  pages={5907--5915},
  year={2017}
}

@inproceedings{zhu2019dm,
  title={DM-GAN: Dynamic memory generative adversarial networks for text-to-image synthesis},
  author={Zhu, Minfeng and Pan, Pingbo and Chen, Wei and Yang, Yi},
  booktitle={Proceedings of the IEEE/CVF Conference on Computer Vision and Pattern Recognition},
  pages={5802--5810},
  year={2019}
}

@inproceedings{Yin_2019_CVPR,
  title={Semantics disentangling for text-to-image generation},
  author={Yin, Guojun and Liu, Bin and Sheng, Lu and Yu, Nenghai and Wang, Xiaogang and Shao, Jing},
  booktitle={Proceedings of the IEEE/CVF Conference on Computer Vision and Pattern Recognition (CVPR)},
  pages={2327--2336},
  year={2019}
}

@inproceedings{szegedy2016rethinking,
  title={Rethinking the inception architecture for computer vision},
  author={Szegedy, Christian and others},
  booktitle={Proceedings of the IEEE/CVF Conference on Computer Vision and Pattern Recognition (CVPR)},
  pages={2818--2826},
  year={2016}
}

@article{schuster1997bidirectional,
  title={Bidirectional recurrent neural networks},
  author={Schuster, Mike and Paliwal, Kuldip K},
  journal={IEEE Transactions on Signal Processing},
  volume={45},
  number={11},
  pages={2673--2681},
  year={1997}
}

@inproceedings{salimans2016improved,
  title={Improved techniques for training GANs},
  author={Salimans, Tim and Goodfellow, Ian and Zaremba, Wojciech and Cheung, Vicki and others},
  booktitle={Advances in Neural Information Processing Systems},
  volume={29},
  year={2016}
}

\end{document}